\documentclass[10pt,twocolumn,letterpaper]{article}

\usepackage{cvpr}
\usepackage{times}
\usepackage{epsfig}
\usepackage{graphicx}
\usepackage{amsmath}
\usepackage{amssymb}

\usepackage{soul}
\usepackage{multirow}
\usepackage{rotating}

\usepackage[pagebackref=true,breaklinks=true,letterpaper=true,colorlinks,bookmarks=false,pdfstartview={FitH}]{hyperref}

\cvprfinalcopy %

\def\mjdraft{1} %

\usepackage[usenames,dvipsnames]{color}
\usepackage{algorithm}
\usepackage[noend]{algpseudocode}

\ifx\mjdraft\undefined

\else

\fi
\newcommand{\myparagraph}[1]{\noindent \textbf{#1}}
\newcommand{\reduceFigSpace}{\vspace{-0.2cm}} %
\newcommand{\OVA}{ `one-vs-all' }

\def\cvprPaperID{450} %

\ifcvprfinal\fi
\begin{document}

\title{Automatic learning of gait signatures for people identification}

\author{F.M. Castro\\
Univ. of Malaga\\
{\tt\small fcastro<at>uma.es}
\and
M.J. Mar\'in-Jim\'enez\\
Univ. of Cordoba\\
{\tt\small mjmarin<at>uco.es}
\and
N. Guil\\
Univ. of Malaga\\
{\tt\small nguil<at>uma.es}
\and
N. P\'erez de la Blanca\\
Univ. of Granada\\
{\tt\small nicolas<at>ugr.es}
}

\maketitle
\begin{abstract}
This work targets people identification in video based on the way they walk (\ie gait). 
While classical methods typically derive gait signatures from sequences of binary silhouettes, in this work 
we explore the use of convolutional neural networks (CNN) for learning high-level descriptors from low-level motion features (\ie optical flow components).
We carry out a thorough experimental evaluation of the proposed CNN architecture on the challenging TUM-GAID dataset. The experimental results indicate that using spatio-temporal cuboids of optical flow as input data for CNN allows to obtain state-of-the-art results on the gait task with an image resolution eight times lower than the previously reported results (i.e. $80\times 60$ pixels). 
\end{abstract}

\section{Introduction} \label{sec:intro}
The goal of \textit{gait recognition} is to identify people by the way they walk. This type of biometric approach is considered non-invasive, since it is performed at a distance, and does not require the cooperation of the subject that has to be identified, in contrast to other methods as iris- or fingerprint-based approaches. Gait recognition has application in the context of video surveillance, ranging from control access in restricted areas to early detection of persons of interest as, for example, v.i.p. customers in a bank office.

From a computer vision point of view, gait recognition could be seen as a particular case of human action recognition. However, 
gait recognition requires more fine-grained features than action recognition, as differences between different gait styles are usually much more subtle than between common action categories (\eg `high jump' vs. `javelin throw') included in state-of-the-art datasets~\cite{soomro2012ucf101}.

\begin{figure}[th]
\begin{center}
   \includegraphics[width=0.98\linewidth]{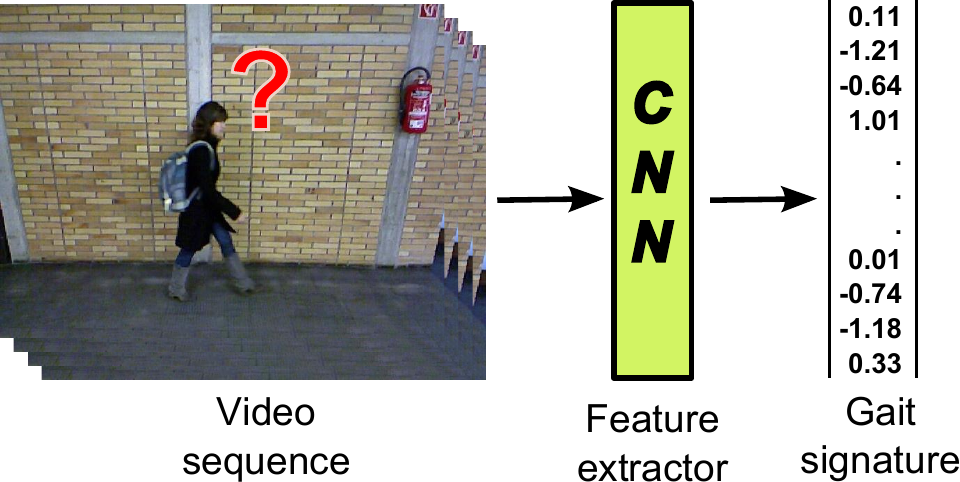}
\end{center}
\caption{\textbf{Goal of this paper}. We aim at automatically learning gait signatures from video sequences of people walking to identify them.}
\label{fig:teaser}
\end{figure}

In last years, great effort has been put into the problem of people identification based on gait recognition~\cite{hu2004survey}.
However, previous approaches have mostly used hand-crafted features for representing the human gait, which are not easily scalable to diverse datasets. 
Therefore, we propose an end-to-end approach based on convolutional neural networks that given low-level optical flow maps, directly extracted from video frames (see Fig.~\ref{fig:teaser}), is able to learn and extract higher-level features suitable for representing human gait: \textit{gait signature}. 

To the best of our knowledge, this is the first work where convolutional neural networks are applied to the problem of gait identification using as input optical flow features. 
Therefore, our main contributions are: 
(i) a preprocessing stage to extract, organize and normalize low-level motion features for defining the input data;
(ii) a convolutional neural network architecture to extract discriminative gait signatures from low-level motion features; and,
(iii) a thorough experimental study to validate the proposed framework on the standard TUM-GAID dataset for gait identification, obtaining state-of-the-art results with video frames whose size is eight times smaller than the ones used in previously reported results.

The rest of the paper is organized as follows. We start by reviewing related work in Sec.~\ref{sec:relwork}. 
An overview of the fundamentals of convolutional neural networks is presented in Sec.~\ref{sec:cnnover}.
Sec.~\ref{sec:approach} explains our approach for learning gait signatures and identifying people. 
Sec.~\ref{sec:expers} contains the experiments and results. 
Finally, we present the conclusions and future work in Sec.\ref{sec:conclu}.

The CNN model obtained for gait recognition is available at:
\href{http://www.uco.es/~in1majim/research/cnngaitof.html}{www.uco.es/\~{}in1majim/research/cnngaitof.html}

\section{Related work}\label{sec:relwork}
\myparagraph{Hand-crafted features.} This is the traditional representation used in gait recognition. Two main approaches stand out over the rest: silhouette-based and dense trajectories-based. 
Silhouette-based descriptors are the most used in the state-of-the-art frameworks. In this sense, the most popular silhouette-based gait descriptor is the called Gait Enery Image (GEI)~\cite{han2006gei}. The key idea is to compute a temporal averaging of the binary silhouette of the target subject. 
To improve the performance of gait recognition, Liu \etal~\cite{liu2012icpr} propose the computation of HOG descriptors from GEI and the Chrono-Gait Image (CGI).
Martin-Felez and Xiang~\cite{martin2014pr}, using GEI as the basic gait descriptor, propose a new ranking model that allows to leverage training data from different datasets.
Hu proposes in \cite{hu2013} the use of a regularized local tensor discriminant analysis method with the Enhanced Gabor representation of the GEI. In addition, the same author defines 
in~\cite{hu2014} a method to identify camera viewpoints at test time from patch distribution features.
Lately, Guan \etal~\cite{guan2015pami} proposed a novel approach to deal with covariate factors (\eg clothing, elapsed time, carrying condition, shoe type) in gait recognition using the GEI descriptor as basis.
Although the use of binary silhouettes is widely extended and has shown excellent results in several scenarios, the computation of noiseless silhouettes is a critical issue, not always easy to achieve. Therefore, in this paper we choose not to use those features.
Recently, an increasing number of publications based on dense trajectories have appeared in the context of action recognition in video. 
The main idea of these approaches is to compute short-term trajectories of densely sampled points for describing, mainly, human motion. Dense trajectories are described with the concatenation of different histograms, like Histograms of Oriented Gradients (HOG), Histograms of Optical Flow (HOF) and Motion Boundary Histograms (MBH) \cite{wang2011cvpr}. An alternative to this representation is the proposed by Jain \etal~\cite{jain2013cvpr} where instead of using HOG, HOF and MBH, they use a new kind of descriptor (Divergence-Curl-Shear) based on partial derivatives of the optical flow. Finally all these trajectories are summarized at video level by using Fisher Vectors~\cite{perronnin2007cvpr} as in \cite{gong2013fisher}. %
A successful gait descriptor based on this approach is the called `Pyramidal Fisher Motion'~\cite{castro2014icpr}, 
which has reported state-of-the-art results on several gait datasets~\cite{castro2015caip,marin2015prl}. 
However, it requires the application of a set of carefully selected feature extraction steps and machine learning techniques, what it is against the goal of this paper.

\myparagraph{Deep-learnt features.} Traditionally, deep learning approaches based in Convolutional Neural Networks (CNN) have been used in image-based tasks with great success \cite{krizhevsky2012nips,simonyan2014corr,zeiler2014eccv}. In the last years, deep architectures for video have appeared, specially focused on action recognition, where the inputs of the CNN are subsequences of stacked frames. In~\cite{simonyan2014nips}, Simonyan and Zisserman proposed to use as input to a CNN a volume obtained as the concatenation of frames with two channels that contain the optical flow in the $x$-axis and $y$-axis respectively. To normalize the size of the inputs, they split the original sequence in subsequences of 10 frames, considering each subsample independently.
Donahue \etal~\cite{donahue2015cvpr} propose another point of view in deep learning using a novel architecture called ``Long-term Recurrent Convolutional Networks''. This new architecture combines CNN (specialized in spatial learning) with Recurrent Neural Networks (specialized in temporal learning) to obtain a new model able to deal with visual and temporal features at the same time.
Recently, Wang \etal~\cite{wang2015cvpr} combined dense trajectories with deep learning. The idea is to obtain a powerful model that combines the deep-learnt features with the temporal information of the trajectories. 
They train a traditional CNN and use dense trajectories to extract the deep features to build a final descriptor that combines the deep information over time. 
On the other hand, Perronnin \etal~\cite{perronnin2015cvpr} propose a more traditional approach using Fisher Vectors as input to a Deep Neural Network instead of using other classifiers like SVM.
Although several papers can be found for the task of human action recognition using deep learning techniques, it is hard to find such type of approaches applied to the problem of gait recognition. In \cite{hossain2013}, Hossain and Chetty propose the use of Restricted Boltzmann Machines to extract gait features from binary silhouettes, but a very small probe set (i.e. only ten different subjects) were used for validating their approach.
Our approach takes the idea of Simonyan and Zisserman~\cite{simonyan2014nips} and uses a spatio-temporal volume of optical flow as input to a CNN specially designed for gait recognition. 
\begin{figure*}[t]
\begin{center}
   \includegraphics[width=0.98\textwidth]{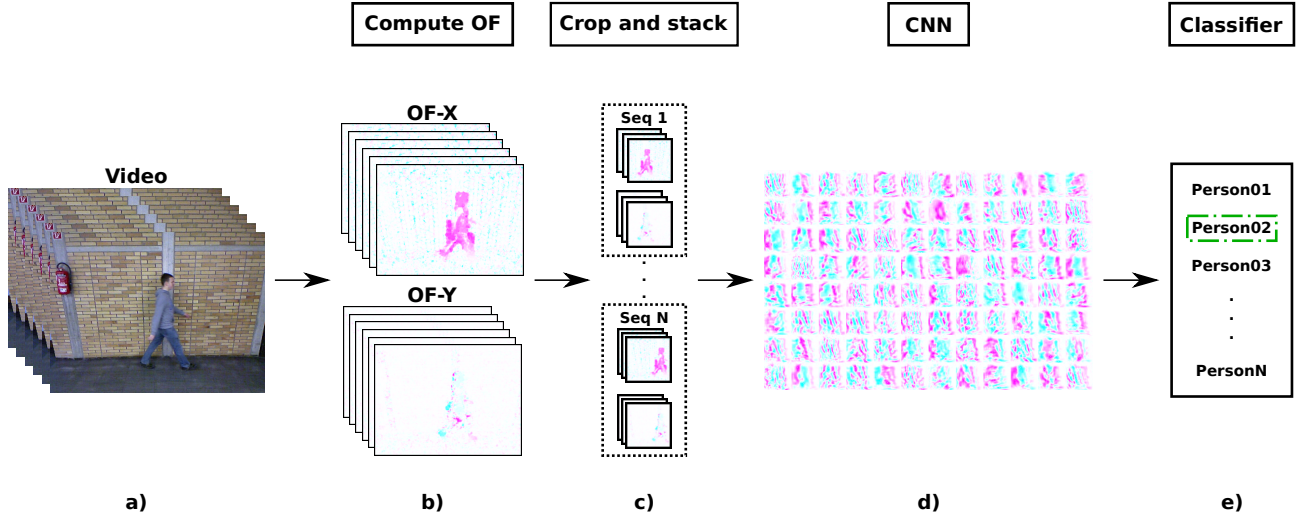}
\end{center}
\reduceFigSpace
\caption{\textbf{Pipeline for gait recognition}. 
a) The input is a sequence of RGB video frames. b) Optical flow is computed along the sequence. c) Optical flow maps are cropped and stacked in subsequences of $L$ maps. d) Optical flow subsequences are passed through the CNN to obtain gait signatures. e) Classification of the extracted gait signatures. Note: positive flows are displayed in pink and negative flows in blue (best viewed in color).}
\label{fig:pipeline}
\end{figure*}
\section{CNN overview} \label{sec:cnnover}
The convolutional neural network (CNN) model is an important type of feed-forward neural network 
with special success on applications where the target information can be represented by a hierarchy of local features (see \cite{bengio2015book}). 
A CNN is defined as the composition of several convolutional layers and several fully connected layers. Each convolutional layer is, in general, the composition of a non-linear layer and a pooling or sub-sampling layer to get some spatial invariance.  For images, the non-lineal layer of the CNN takes advantage, through local connections  and weight sharing, of the 2D structure present in the data. These two conditions impose a very strong regularization on the total number of weights in the model, which allows a successful training of the model by using back-propagation. In our approach, although we do not feed the model directly with the RGB image pixels, the CNN approach remains relevant since the optical flow information also shares the local dependency property as the pixels do.

In the last years, CNN models are achieving state-of-the-art results on many different complex applications (\eg object detection, text classification, natural language processing, scene labeling, etc.) \cite{collobert2011,krizhevsky2012nips,farabet2013,zhang2015}. 
However, to the extent of our knowledge, CNN has not been applied to the problem of gait recognition yet.
The great success of the CNN model is in part due to 
its use on data where the target can be represented through a feature hierarchy of increasing semantic complexity. When a CNN is successfully trained, the output of the last hidden layer can be seen as the coordinates of the %
target in a high level representation space. The fully connected layers, on top of the convolutional ones, allow us to reduce the dimensionality of such representation and, therefore, to improve the classification accuracy.

\section{Proposed approach} \label{sec:approach}
In this section we describe our proposed framework to address the problem of gait recognition using CNN. 
The pipeline proposed for gait recognition based on CNN is represented in Fig.~\ref{fig:pipeline}: 
\textit{(i)} compute optical flow (OF) along the whole sequence; 
\textit{(ii)} build up a data cuboid from consecutive OF maps; 
\textit{(iii)} feed the CNN with OF cuboid to extract the gait signature; and, 
\textit{(iv)} apply a classifier to decide the subject identity.

\subsection{Input data} \label{subsec:features}
The use of optical flow (OF) as input data for action representation in video with CNN has already shown excellent results~\cite{simonyan2014nips}.
Nevertheless human action is represented by a wide, and usually well defined, set of local motions. In our case, the set of motions differentiating one gait style from another is much more subtle and local. An important question here is whether the gait information can be decoded from simple and low resolution (\eg $80 \times 60$) optical flow.

Let $F_t$ be an OF map computed at time $t$ and, therefore, $F_t(x,y,c)$ be the value of the OF vector component $c$ located at coordinates $(x,y)$, where $c$ can be either the horizontal or vertical component of the corresponding OF vector. 
The input data $I_L$ for the CNN are cuboids built by stacking $L$ consecutive OF maps $F_t$, where $I_L(x,y,2k-1)$ and $I_L(x,y,2k)$ corresponds to the value of the horizontal and vertical OF components located at spatial position $(x,y)$ and time $k$, respectively, ranging $k$ in the interval $[1,L]$.

\begin{figure}[t]
\begin{center}
   \includegraphics[width=0.98\linewidth]{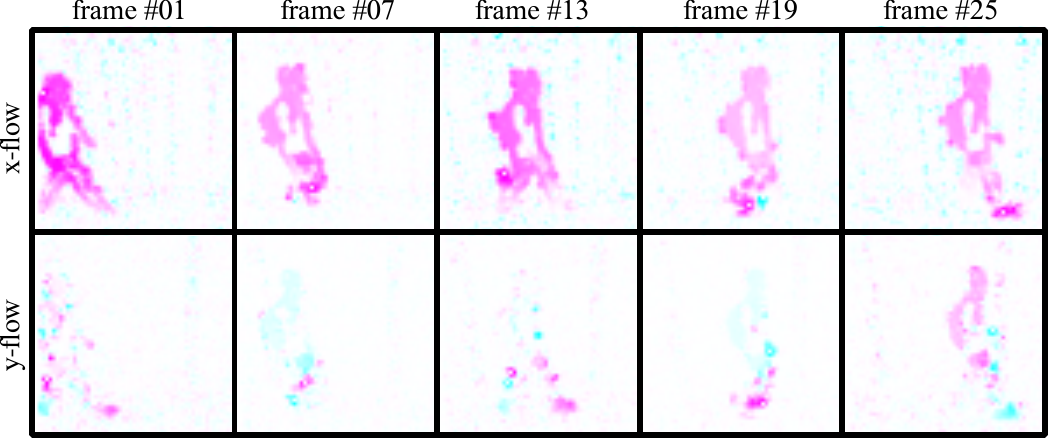}
\end{center}
\caption{\textbf{Input data}. Frames extracted from a subsequence of $25$ frames. (top) Optical flow in $x$-axis. (bottom) Optical flow in $y$-axis. Note: positive flows are displayed in pink and negative flows in blue (best viewed in color). }
\label{fig:subseqs}
\end{figure}

Since each original video sequence will probably have a different temporal length, and CNN requires a fixed size input, we extract subsequences of $L$ frames from the full-length sequences. 
In Fig.~\ref{fig:subseqs} we show five frames distributed every six frames along a subsequence of twenty-five frames in total (i.e. frames 1, 7, 13, 19, 25).
Top row frames show the horizontal component of the OF ($x$-axis displacement) and bottom row frames show the vertical component of the OF ($y$-axis displacement). It can be observed that most of the flow is concentrated in the horizontal component, due to the displacement of the person. 
In order to remove noisy OF located in the background, as it can be  observed in Fig.~\ref{fig:subseqs}, we might think in applying a preprocessing step for filtering out those vectors whose magnitude is out of a given interval. However, since our goal in this work is to minimize the manual intervention in the process of gait signature extraction, we will use those OF maps as returned by the OF algorithm.

\myparagraph{Implementation details} \label{subsec:implemOF}
First of all, we resize the RGB video frames to a common size of $80 \times 60$ pixels, keeping the original aspect ratio of the video frames. Then, 
we compute dense OF on pairs of frames by using the method of Farneback~\cite{Farneback03} implemented in OpenCV library. In parallel, people are located in a rough manner along the video sequences by background substraction~\cite{kaewtrakulpong2002bmm}. 
Then, we crop the video frames to remove part of the background, obtaining video frames of $60\times 60$ pixels (full height is kept) and to align the subsequences (people are $x$-located in the middle of the central frame, \#13) as in Fig.~\ref{fig:subseqs}.

Finally, from the cropped OF maps, we build subsequences of $25$ frames by stacking OF maps with an overlap of $\mathcal{O}\%$ frames. In our case, we chose $\mathcal{O}=80\%$, that is, to build a new subsequence, we use $20$ frames of the previous subsequence and $5$ new frames. For most state-of-the-start datasets, 25 frames cover almost one complete gait cycle, as stated by other authors~\cite{barnich2009prl}.

In order to increase the number of samples available for training, we compute 8 spatial displacements of
$\pm 5$ pixels in all directions, \ie (-5,-5), (-5, 0), (0,-5), ..., (0,5). Then, the corresponding mirror sequences are computed. These procedure allows us to obtain about $270k$ training samples.
Finally, before feeding each sample into the CNN, the mean value of the whole training dataset is subtracted.

\subsection{CNN architecture for gait signature extraction} \label{subsec:cnnarch}
The CNN architecture we propose for gait recognition is based on the one described in \cite{simonyan2014nips} for general action recognition in video. 
However, in our case, the input has a size of $60\times 60 \times 50$, obtained from the sequence of 25 OF frames with their corresponding two channels, as explained in the previous section.

\begin{figure}[t]
\begin{center}
   \includegraphics[width=0.98\linewidth]{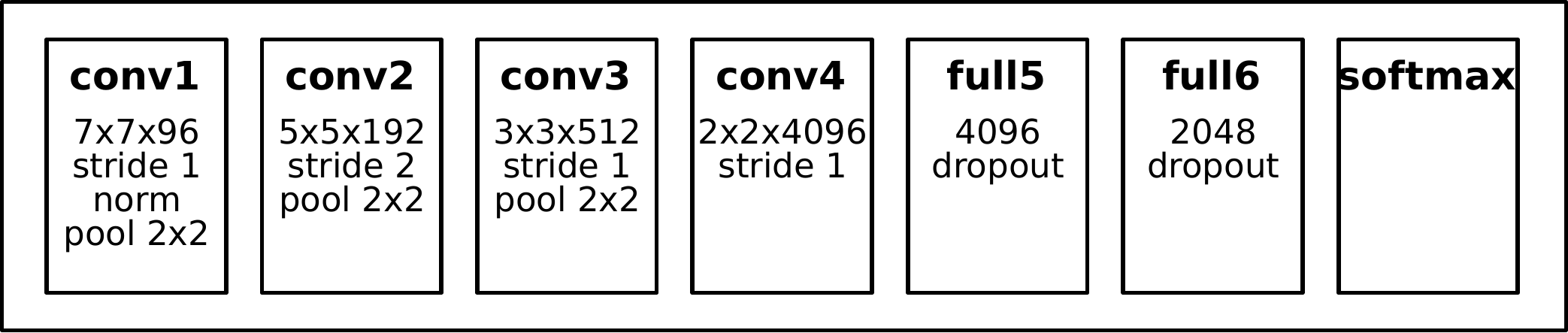}
\end{center}
\caption{\textbf{Proposed CNN architecture for gait signature extraction}. Four convolutional layers are followed by two fully connected layers. The top layer is a softmax classifier that can be used to directly derive an identity.}
\label{fig:archCNN}
\end{figure}

The proposed CNN is composed by the following sequence of layers (Fig.~\ref{fig:archCNN}): 
`\textit{conv1}', 96 filters of size $7\times 7$ applied with stride 1 followed by a normalization and max pooling $2\times 2$;
`\textit{conv2}', 192 filters of size $5\times 5$ applied with stride 2 followed by max pooling $2\times 2$;
`\textit{conv3}', 512 filters of size $3\times 3$ applied with stride 1 followed by max pooling $2\times2$;
`\textit{conv4}', 4096 filters of size $2\times 2$ applied with stride 1;
`\textit{full5}', fully-connected layer with 4096 units and dropout;
`\textit{full6}', fully-connected layer with 2048 units and dropout; and,
`\textit{softmax}', softmax layer with as many units as subject identities.
All convolutional layers use the rectification (ReLU) activation function.
\myparagraph{Implementation details} \label{subsec:implemCNN}
We use the implementation of CNN provided in MatConvNet library~\cite{vedaldi2015matconvnet}. This library allows to develop CNN architectures in an easy and fast manner using the Matlab environment. In addition, it takes advantage of CUDA and cuDNN~\cite{chetlur2014cudnn} to improve the performance of the algorithms.

We perform CNN training following an iterative process to speed up and to facilitate the convergence. In this iterative process, initially, we train a simplified version of our CNN 
(i.e. `conv1' without normalization, `conv4' 512 filters, `full5' 512 units, `full6' 256 units and no dropout) and, 
then, we use its weights for initializing the layers of a more complex version of that simpler CNN (i.e. adding normalization, 0.1 dropout and more filters and units). By this way, we train four incremental CNN versions using the previous weights until we obtain the final CNN architecture represented in Fig.~\ref{fig:archCNN}.
During the training of the CNN, the weights are learnt using mini-batch stochastic descent algorithm with momentum equal to $0.9$ in the first three CNN version iterations, and 0.95 during the last one. We set weight decay to $5 \cdot 10^{-4}$ and dropout to $0.4$. The learning rate is initially set to $10^{-2}$ and divided by $10$ when the validation error become stagnant. At each epoch, a mini-batch of 150 samples is constructed by random selection over a balanced training set (i.e. almost same proportion of samples per class).

\subsection{Classification strategies} \label{subsec:classifiers}
Once we have obtained the gait signatures, the final stage consists in classifying those signatures to derive a subject identity.
Although the softmax layer of the CNN is already a classifier (i.e. each unit represents the probability of belonging to a class), the fully-connected layers can play the role of gait signatures that can be used as input of a Support Vector Machine (SVM) classifier. 
Since we are dealing with a multiclass problem, we define an ensemble of $C$ binary SVM classifiers with linear kernel in an \OVA fashion, where $C$ is the number of possible subject identities.
Previous works (\eg \cite{castro2014icpr}) indicate that this configuration of binary classifiers is suitable to obtain top-tier results in this problem.
Note that we $L2$-normalize the top fully-connected layer before using it as feature vector.

A classical alternative to discriminative classifiers is the nearest neighbour (NN) classifier, which does not require any training step. Actually, we can easily extend our gait recognition system by just adding samples of the new subjects to our \textit{gallery set} (i.e. the models).

Note that in Sec.~\ref{subsec:features}, we split the whole video sequence into overlapping subsequences of a fixed length, and those subsequences are classified independently. Therefore, in order to derive a final identity for the subject walking along the whole sequence, we apply a \textit{majority voting} strategy on the labels assigned to each subsequence.

\section{Experiments and results} \label{sec:expers}
We present here the experiments designed to validate our approach and the results obtained on the selected dataset for gait recognition.
\subsection{Dataset} \label{subsec:datasets}
\begin{figure}[t]
\begin{center}
   \includegraphics[width=0.98\linewidth]{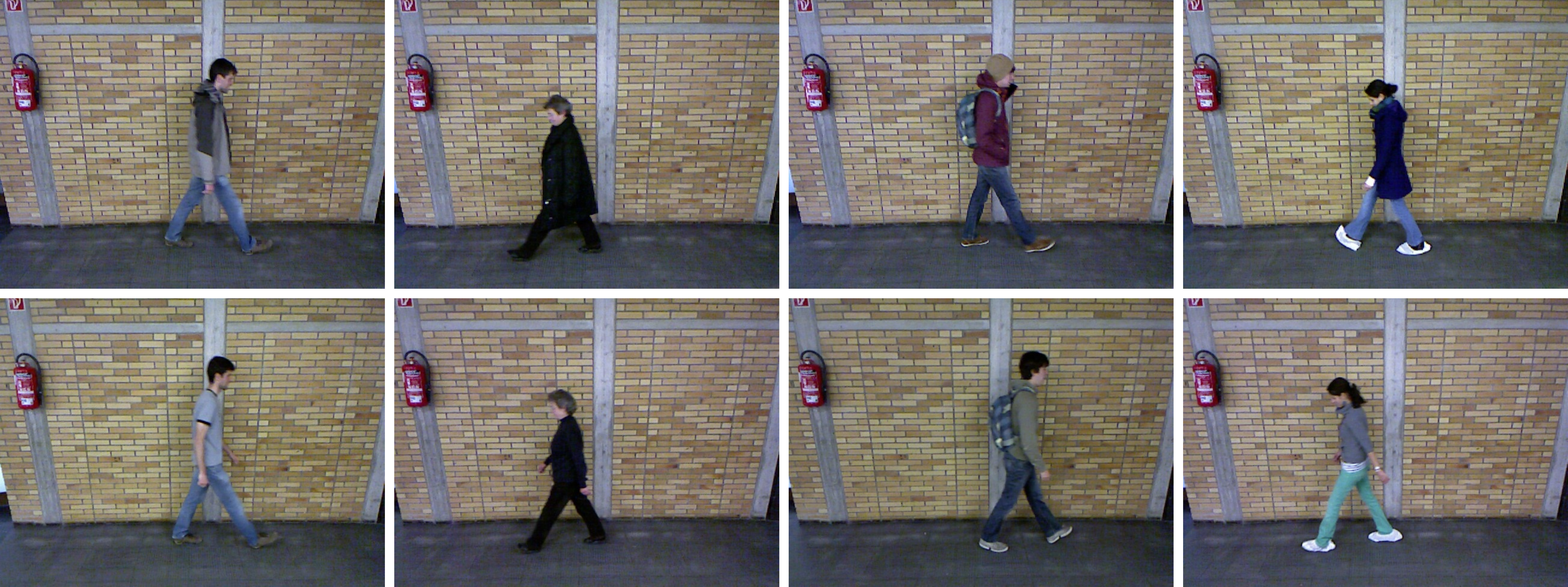}
\end{center}
\caption{\textbf{TUM-GAID dataset}. People walking indoors under four walking conditions: normal walking, wearing coats, carrying a bag and wearing coating shoes. Top and bottom rows show the same set of subjects but in different months of the same year.}
\label{fig:datasets}
\end{figure}

We run our experiments on the recent `TUM Gait from Audio, Image and Depth' (TUM-GAID) dataset~\cite{hofmann2014tumgaid} for gait recognition.
In TUM-GAID 305 subjects perform two walking trajectories in an indoor environment. %
The first trajectory is performed from left to right and the second one from right to left. Therefore, both sides of the subjects are recorded. Two recording sessions were performed, one in January, where subjects wore heavy jackets and mostly winter boots, and the second in April, where subjects wore different clothes. 
The action is captured by a Microsoft Kinect sensor which provides a video stream with a resolution of $640 \times 480$ pixels with a frame rate of approximately 30 fps.
Some examples can be seen in Fig.~\ref{fig:datasets} depicting the different conditions included in the dataset.

Hereinafter the following nomenclature is used to refer each of the four walking conditions considered: \textit{normal} walk (\textit{N}), carrying a \textit{backpack} of approximately 5 kg (\textit{B}), wearing coating \textit{shoes} (\textit{S}), as used in clean rooms for hygiene conditions, 
and \textit{elapsed time} (\textit{TN-TB-TS}). 
Each subject of the dataset is composed of: six sequences of normal walking (\textit{N1, N2, N3, N4, N5, N6}), two sequences carrying a bag (\textit{B1, B2}) and two sequences wearing coating shoes (\textit{S1, S2}). In addition, 32 subjects were recorded in both sessions  (i.e. January and April) so they have 10 additional sequences (\textit{TN1, TN2, TN3, TN4, TN5, TN6, TB1, TB2, TS1, TS2}). Therefore, the overall amount of videos is 3400.

To standardize the experiments performed on the dataset, the authors have defined three subsets of subjects: training, validation and testing. The training set is used for obtaining a robust model against the different covariates of the dataset. This partition is composed of 100 subjects and the sequences \textit{N1} to \textit{N6}, \textit{B1}, \textit{B2}, \textit{S1} and \textit{S2}. The validation set is used for validation purposes and contains 50 different subjects with the sequences \textit{N1} to \textit{N6}, \textit{B1}, \textit{B2}, \textit{S1} and \textit{S2}. Finally, the test set contains other 155 different subjects used in the test phase. As the set of subjects is different between the test set and the training set, a new training of the identification model must be performed. 
For this purpose, the authors reserve the sequences \textit{N1} to \textit{N4}, from the subject test set, to train the model again and the rest of sequences are used for testing and to obtain the accuracy of the model. In the \textit{elapsed time} experiment, the temporal sequences (\textit{TN1, TN2, TN3, TN4, TN5, TN6, TB1, TB2, TS1, TS2}) are used instead of the normal ones and the subsets are: 10 subjects in the training set, 6 subjects in the validation set and 16 subjects in the test set.

For the viability of our experiments with CNN, we resized all the videos to a resolution of $80 \times 60$ pixels (i.e. 8 times lower resolution). Nevertheless, we will show in the experimental results (Sec.~\ref{subsec:results}), that we obtain state-of-the-art results with such low resolution, what, in our opinion, highlights the potential of CNN for gait recognition.

\subsection{Performance evaluation} \label{subsec:metrics}
For each test sample, we return a sorted list of possible identities, where the top one identity corresponds to the largest scored one. Therefore, we use the following metrics to quantitative measure the performance of the proposed system: \textit{rank-1} and \textit{rank-5}. 
Metric \textit{rank-1} measures the percentage of test samples where the top one assigned identity corresponds to the right one. Whereas \textit{rank-5} measures the percentage of test samples where the ground truth identity is included in the first five ranked identities for the corresponding test sample. Note that \textit{rank-5} is less strict than \textit{rank-1} and, in a real system, it would allow to verify if the target subject is any of the top 5 most probably ones.

\subsection{Experimental setup} \label{subsec:expersetup}
We describe here the experiments we carried out on the dataset with the proposed approach.

\myparagraph{Experiment A: gait recognition with clothing and carrying conditions.}
This is the core experiment of this paper, where we aim at evaluating the capacity of the proposed CNN model to extract gait signatures robust enough to deal with covariate factors as clothing changes (\eg. long coats or coating shoes) or carrying conditions (\eg backpacks). In fact, the CNN model trained here will be used for the subsequent experiments.

Training of the CNN convolutional filters is carried out by using only sequences of the standard training and validation subject partitions (i.e. $100+50$ subjects) of TUM-GAID, including the three scenarios. 
Once the CNN model is trained with those samples, the learnt weights from layers `conv1' to `full6' are frozen (i.e. not modified any more). 
In order to evaluate the performance of the CNN-based gait signatures on the test subject partition (i.e. $155$ subjects),  %
only the softmax layer will be fine-tuned by using the training sequences of scenario `N' from the test subject partition, as the subject identities have changed. However, when we use SVM or NN classifiers, no CNN fine-tuning is needed, as we will use the output of layer `full6' directly as our gait signature (i.e. the automatic gait descriptor extracted from the input sequence).

The results of this experiment are summarized in Tab.~\ref{tab:experA},  
where each row corresponds to a different combination of features and classifiers: softmax `SM', support vector machine `SVM' and nearest neighbour `NN'. Each column contains the recognition results of the diverse scenarios included in the dataset (\textit{N}, \textit{B}, \textit{S}) plus the average on the three scenarios (`Avg'). For completeness, we report rank-1 (`R1') and rank-5 (`R5') results.

Moreover, for comparison purposes, we have implemented the `Pyramidal Fisher Motion' (PFM) descriptor, as described in~\cite{castro2014icpr}, since it does not need binary silhouettes as input for its computation and has previously reported state-of-the-art results for the problem of gait recognition~\cite{castro2015caip}. 
Note that we have used the PFM descriptor both in the original resolution video sequences (row `PFM@$640\times 480$') and in the low resolution version of the sequences (row `PFM@$80\times 60$'), to allow a fair comparison with our CNN-based gait signatures that use the low resolution version. 
For `PFM@$640\times 480$', we have used the whole video sequence to compute a single descriptor, as in the original paper~\cite{castro2014icpr}. Whereas in `PFM@$80\times 60$', we have computed several PFM using the same set of subsequences extracted for CNN, making an even much fairer comparison. After the classification of each PFM of the sequence, majority voting was applied to obtain a final identity.

\myparagraph{Experiment B: elapsed time.} \label{subsec:elapsed}
The goal of this experiment is to evaluate the 
robustness of the CNN-based gait signatures against changes of people appearance at different periods of time.
In this experiment, we apply the CNN model trained in `Experiment A' on the `elapsed time' subset of TUM-GAID (Sec.~\ref{subsec:datasets}), which is composed of 16 subjects for training and validation, and 16 for testing.  
From the training sequences of the `normal' scenario \textit{TN} of the 16 test subjects, we obtained 10620 samples that were used to fine-tune the softmax layer of the CNN trained in the previous experiment, as the subject identities changed.
Then, we used the test sequences of the three \textit{elapsed time} scenarios to evaluate the performance.

The results of this experiment are summarized in Tab.~\ref{tab:experB}, where each row corresponds to a different combination of features and classifiers, including PFM. Each column presents the recognition results of the diverse scenarios included in the \textit{elapsed time} subset (\textit{TN}, \textit{TB}, \textit{TS}) plus the average on the three scenarios (`Avg'). For completeness, we report rank-1 (`R1') and rank-5 (`R5') results.

\myparagraph{Experiment C: gait-based gender recognition.} \label{subsec:gender}
Gender recognition based on gait signatures is considered a kind of soft biometric, which allows to prune a subset of subjects for a subsequent finer identification.
The goal of this experiment is to validate the quality of the gait signatures learnt in the first experiment to train a binary linear SVM for gender classification. For evaluation purposes, we train the gender classifier only on the gait sequences included in the training and validation subject partitions. 
In TUM-GAID, which provides labels at video level for this task, the proportion of male and female subjects in the test set is 62.6\% and 37.4\%, respectively.

The results of this experiment are summarized in Tab.~\ref{tab:experGender}, where we show both the confusion matrices for each scenario, plus the overall accuracy of the classifier. For comparison purposes, bottom row contains the accuracy reported for this task in paper~\cite{hofmann2014tumgaid}.

\begin{figure}[t]
\begin{center}
   \includegraphics[width=0.98\linewidth]{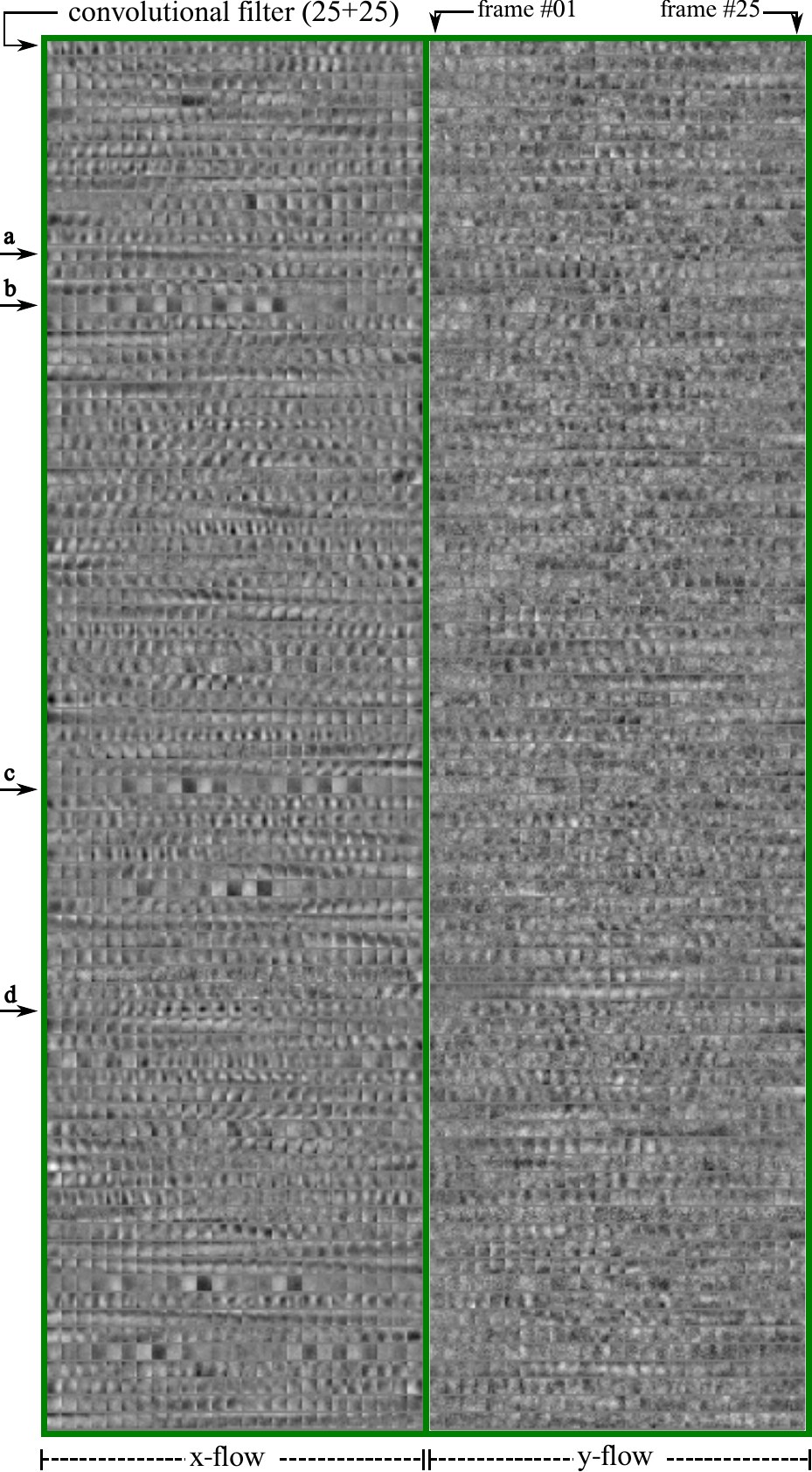}
\end{center}
\reduceFigSpace
\caption{\textbf{Learnt filters}. 96 convolutional filters of the first CNN layer. Each row contains two filters corresponding to $x$-axis (left) and $y$-axis (right). Each column contains a channel of the filter (i.e. each frame of the subsequence). (Best viewed in digital format)}
\label{fig:filters}
\vspace{-0.4cm}
\end{figure}
\subsection{Results and discussion} \label{subsec:results}
We ran our experiments on a computer with 32 cores at 2 GHz, 256 GB of RAM and a GPU Nvidia Tesla K40c, %
with MatConvNet library running on Matlab 2014b for Ubuntu 14.04. 
After splitting the training sequences (of the training subjects) into subsequences, we got a training set composed of 269352 samples used for learning the filters from `conv1' to `full6' layers (see Fig.~\ref{fig:archCNN}); and a second training set composed of 108522 samples for training the softmax layer from the subset of test subjects\footnote{Note that TUM-GAID distinguishes between training/test \textit{subjects} and training/test \textit{sequences}. Test sequences are never used for training or validation of the model.}. 
With all these samples, the whole training process (from the first CNN model until the fine-tuning of the softmax layer of the final model) took about 60 hours.

Due to the specificity of the dataset, in the training step we had to balance the number of samples of the different kinds of walking (i.e. normal, carrying a bag and wearing coating shoes). To do this, we defined different training subsets with the same number of samples of each walking scenario, and when the CNN converged, we continued the training with a different subset. At the end of the training phase, all samples of the original training set had been passed through the CNN at least twice to guarantee a good performance of the model. If this step is not performed, the CNN would learn mainly specialized filters for `normal' walk, as we have four times more samples of this kind than the others.

We show in Fig.~\ref{fig:filters} the 96 convolutional filters learnt at the first CNN layer during training for `Experiment A'. Each row shows a pair of set of filters corresponding to the horizontal (left) and vertical (right) components of the OF. Each represented filter component spans $7\times 7$ pixels.
The first aspect that is appreciable in this set of filters is that there seems to be two main types of filters:
filters acting as \textit{spatial derivatives}, where patterns are distinguishable like in rows `a' and `d' (the pattern evolves along time); 
and, filters acting as \textit{temporal derivatives}, where the mask in each frame is mainly uniform but changes its intensity along frames like in rows `b' and `c'. These observations are shared with the ones made by Simonyan and Zisserman in~\cite{simonyan2014nips} applied to action datasets.
The second aspect that we can perceive is the difference between $x$-flow filters and $y$-flow filters. The set of $x$-flow filters exhibits a structure more defined than the $y$-flow filters, which are more noisy and blurry. In our opinion, this difference is due to the fact that the main motion in the gait is located in the horizontal axis, as the displacement of the subject is along such axis. In contrast, vertical movements (i.e. body limbs) are softer and subtler, getting filters less defined. %

\begin{table}
\small
\begin{center}
\setlength{\tabcolsep}{0.25em} %
\begin{tabular}{|l|c c|c c|c c||c c|}
\hline
  & \multicolumn{2}{c|}{\textit{N}} & \multicolumn{2}{c|}{\textit{B}} & \multicolumn{2}{c||}{\textit{S}} & \multicolumn{2}{c|}{\textit{Avg}}\\
\hline
 \textit{Method} / \textit{Rank} & R1 & R5 & R1 & R5 & R1 & R5 & R1 & R5 \\
\hline
CNN-SM & 99.4 & 100 & 94.5 & 99.4 & 94.2 & 98.7 & 96.0 & 99.4 \\ %
CNN-SVM & 99.7 & 100 & 97.1 & 99.4 & 97.1 & 99.4 & \textbf{98.0} & \textbf{99.6} \\
\hline
CNN-NN+PCA256 & 99.4 & 99.7 & 97.7& 98.7 & 96.1& 97.7 & 97.7 & 98.7\\ %
CNN-NN+PCA128 & 99.7 & 100 & 98.1& 98.4 & 95.8&97.1 &97.9 & 98.5 \\ %
CNN-NN+PCA064 & 99.7 & 100 & 98.1& 98.4 & 94.8 & 96.4 &97.5 & 98.3\\ %
\hline
PFM@$80 \times 60$  & 75.8 & 93.2 & 70.3 & 91.3 & 32.3 & 62.9 & 59.5 & 82.5 \\  %
PFM@$640\times 480$ & 99.7 & 99.7 & 99.0 & 99.4 & 99.0 & 99.4 & \textbf{99.2} & \textbf{99.5} \\ %
\hline
\end{tabular}
\end{center}
\caption{\textbf{Experiment A. } Percentage of correct recognition on scenarios \textit{N-B-S} of TUM-GAID dataset by using \textit{rank-1} (R1) and \textit{rank-5} (R5) metrics. Each row corresponds to a different combination of features and classifiers. Best average results are marked in bold.} 
\label{tab:experA}
\end{table}
Focusing on `Experiment A', the results in Tab.~\ref{tab:experA} indicate that from low resolution frames (i.e. $80 \times 60$) the trained CNN model is able to extract gait signatures that used in combination with standard SVM classifiers, it is attained an average of $98\%$ rank-1 correct recognition (see row `CNN-SVM'), and $99.6\%$ of rank-5 accuracy. 
Comparing SVM with SM, we can see that the obtained results are quite similar, although SVM accuracy is slightly better, indicating a good linear separability of the test subjects given the extracted gait signatures.
For speeding-up the NN classifier, the 2048-dimensional gait descriptors were compressed with the standard principal components analysis (PCA) algorithm -- vectors are $L2$-normalized and mean is subtracted before PCA -- obtaining compact signatures of 64, 128 and 256 dimensions. The average results reported in rows `CNN-NN+PCA\textit{x}' are comparable to the ones yielded by the parametric classifiers (i.e. SVM and SM),  making attractive the use of NN in combination with these CNN-based signatures as no training stage is needed if adding new identities to our recognition system is required.
Furthermore, our proposal outperforms PFM descriptor when used on the same low resolution video sequences (i.e. $59.5\%$ vs. $98\%$), although average rank-1 accuracy for PFM at full resolution is around $1\%$ better that CNN. Nevertheless, our CNN-based signature extractor has been trained in a fully automatic manner, in contrast to the hand-crafted steps need for computing PFM.
Focusing on the results on scenarios `B' and `S', we can conclude that our CNN signatures are able to successfully represent the discriminative motion patterns that characterize the different subjects regardless the clothing or shoes worn or the bags carried. 
Remember that we used a set of individuals totally different for training the CNN filters than the one used for testing the signatures obtained with them.

\begin{table}
\small
\begin{center}
\setlength{\tabcolsep}{0.25em} %
\begin{tabular}{|l|c c|c c|c c||c c|}
\hline
  & \multicolumn{2}{c|}{\textit{TN}} & \multicolumn{2}{c|}{\textit{TB}} & \multicolumn{2}{c||}{\textit{TS}} & \multicolumn{2}{c|}{\textit{Avg}}\\
\hline
 \textit{Method} / \textit{Rank} & R1 & R5 & R1 & R5 & R1 & R5 & R1 & R5 \\
\hline
CNN-SM & 53.1 & 87.5 & 40.6 & 90.6 & 50.0 & 90.6 & 47.9 & \textbf{89.6} \\ %
CNN-SVM & 59.4 & 87.5 & 50.0 & 78.1 & 62.5 & 93.8 & 57.3 & 86.5 \\
\hline
CNN-NN+PCA256 & 59.4 & 71.9 & 56.3 & 65.6 & 56.3 & 65.6 & 57.3 & 67.7\\ %
CNN-NN+PCA128 & 62.5 & 71.9 & 56.3 & 65.6 & 59.4 & 68.8 & \textbf{59.4} & 68.8\\ %
CNN-NN+PCA064 & 62.5 & 68.8 & 53.1 & 62.5 & 59.4 & 68.8 & 58.3 & 66.7\\ %
\hline
PFM@$80 \times 60$  & 50.0 & 84.4 & 40.6 & 81.3 & 25.0 & 75.0 & 38.5 & 80.2 \\  %
PFM@$640\times 480$ & 78.1 & 87.5 & 56.3 & 87.5 & 46.9 & 87.5 & \textbf{60.4} & 87.5 \\
\hline
\end{tabular}
\end{center}
\caption{\textbf{Experiment B. } Percentage of correct recognition on scenarios \textit{TN-TB-TS} of TUM-GAID dataset by using \textit{rank-1} (R1) and \textit{rank-5} (R5) metrics. Each row corresponds to a different combination of features and classifiers. Best average results are marked in bold.} 
\label{tab:experB}
\end{table}
Moving to `Experiment B', the `elapsed time' experiment proposed in TUM-GAID is more challenging than the previous one, as there is a temporal gap of months between recordings of the same subjects. This higher level of difficulty is reflected in the results of Tab.~\ref{tab:experB}, where we directly use the CNN previously trained in `Experiment A'. In terms of rank-1 accuracy, SVM behaves on average better than SM, as previously reflected in `Experiment A'. However, NN classifiers improves on SVM, what suggests that this set of subjects is not linearly separable given the gait signatures. 
Comparing to PFM, only the full resolution version obtains average results slightly better than CNN, due to the good results achieved in the normal (\textit{TN}) scenario. Note that results reported for \textit{TB} and \textit{TS} are equal or lower than the CNN ones.
\begin{table}[t] %
\centering
\small
\setlength{\tabcolsep}{0.25em} %
\begin{tabular}{|c|c|ccc|c||ccc|c|}
\hline 
&\textit{Method} & \textit{N} & \textit{B} & \textit{S} & \textit{Avg} & \textit{TN} & \textit{TB} & \textit{TS} & \textit{Avg}\\ 
\hline 
\multirow{7}{*}{\rotatebox{90}{$640\times 480$}} &SDL~\cite{zeng2014pr} & - & - & - & - & \textbf{96.9} & - & - &  - \\ 
&GEI~\cite{hofmann2014tumgaid} & 99.4 & 27.1 & 52.6 & 59.7 & 44.0 & 6.0 & 9.0 & 19.7 \\
&SEIM~\cite{whytock2014jmiv} & 99.0 & 18.4 & 96.1 & 71.2 & 15.6 & 3.1 & 28.1 & 15.6 \\
&GVI~\cite{whytock2014jmiv} & 99.0 & 47.7 & 94.5 & 80.4 & 62.5 & 15.6 & \textbf{62.5} & 46.9 \\
&SVIM~\cite{whytock2014jmiv} & 98.4 & 64.2 & 91.6 & 84.7 & 65.6 & 31.3 & 50.0 & 49.0 \\
&RSM~\cite{guan2013icb} & \textbf{100.0} & 79.0 & 97.0 & 92.0 & 58.0 & 38.0 & 57.0 & 51.3\\
&PFM~\cite{castro2014icpr} & 99.7 & \textbf{99.0} & \textbf{99.0} & \textbf{99.2} & 78.1 & \textbf{56.3} & 46.9 & \textbf{60.4}\\
\hline 
\multirow{1}{*}{\tiny $80\times 60$}&CNN-SVM & 99.7 & 97.1 & 97.1 & 98.0 & 59.4 & 50.0 & \textbf{62.5} & 57.3 \\
\multirow{1}{*}{\tiny (ours)} & {\footnotesize CNN-NN128} & 99.7 & 98.1 & 95.8 & 97.9 & 62.5 & \textbf{56.3} & 59.4 & 59.4 \\
\hline
\end{tabular}
\caption{\textbf{State-of-the-art on TUM GAID}. Percentage of correct recognition on TUM-GAID for diverse methods published in the literature. Bottom row corresponds to our proposal, where instead of using video frames at $640\times 480$, a resolution of $80\times 60$ is used. Each column corresponds to a different scenario. Best results are marked in bold. (See main text for further details). }
\label{tab:tumcomp}
\vspace{-0.4cm}
\end{table}
Comparing our best results with previously published ones, we observe in Tab.~\ref{tab:tumcomp} that our accuracy (rank-1) is on a par with those methods, even though we are using video frames with a resolution eight times lower than the others. Note that our average accuracy (columns `Avg') in both sets of experiments is greater than the ones reported in all the compared papers, but PFM, what emphasizes the quality of the gait signatures returned by the proposed CNN.
\begin{table}
\begin{center}
\scriptsize
\setlength{\tabcolsep}{0.2em} %
\begin{tabular}{|c|r|c c||c c||c c|}
\cline{3-8}
  \multicolumn{1}{c}{} & & \multicolumn{2}{c||}{\textit{N}} & \multicolumn{2}{c||}{\textit{B}} & \multicolumn{2}{c|}{\textit{S}}\\
\hline
\multirow{3}{*}{\rotatebox{90}{\textit{CM}}}  & & Female & Male & Female & Male & Female & Male \\
\cline{3-8} %
& Female & \textbf{78.4} & 21.6 & \textbf{77.6} & 22.4 & \textbf{76.7} & 23.3 \\
& Male & 4.6 & \textbf{95.4} & 4.6 & \textbf{95.4} & 3.6 & \textbf{96.4} \\
\hline \hline
\multirow{2}{*}{\rotatebox{90}{\textit{Acc}}} & CNN+SVM (ours, $80 \times 60$)    & \multicolumn{2}{c||}{89.0\%} & \multicolumn{2}{c||}{88.7\%} & \multicolumn{2}{c|}{89.0\%} \\

&Hofmann \etal~\cite{hofmann2014tumgaid} ($640 \times 480$)& \multicolumn{2}{c||}{95.8\%} & \multicolumn{2}{c||}{74.8\%} & \multicolumn{2}{c|}{92.9\%} \\
\hline
\end{tabular}
\end{center}
\caption{\textbf{Experiment C. } Confusion matrices for gender recognition based on convolutional gait signatures. For each scenario, each cell in `CM' contains the percentage of probe samples assigned to each gender.
Rows in `Acc' contain the overall accuracy per scenario.} 
\label{tab:experGender}
\vspace{-0.4cm}
\end{table}
Finally, the results in Tab.~\ref{tab:experGender} suggest that the problem of gender recognition can be successfully addressed based on just motion features (i.e. optical flow). Regardless the scenario, accuracy for female recognition is lower than male recognition (i.e. $\approx 77\%$ vs. $\approx 96\%$). This can be due to the ratio among female and male samples in the dataset.
Comparing to the results reported by Hofmann \etal in \cite{hofmann2014tumgaid}, the average on the three scenarios (\textit{N}, \textit{B}, \textit{S}) for our method is $88.9\%$, whereas the average for their method is $87.8\%$, despite the lower resolution of our video inputs.
\section{Conclusions and future work} \label{sec:conclu}
This paper has presented a thorough study of convolutional neural networks applied to the demanding problem of people identification based on gait.
The experimental validation has been carried out on the challenging dataset TUM-GAID, by using a low resolution version of the original video sequences (i.e. eight times lower). 
The results indicate that starting from just sequences of optical flow, the proposed CNN is able to extract meaningful gait signatures (i.e. $L2$-normalized top fully-connected layer) that allow to obtain high recognition rates on the available scenarios (i.e. different clothing and wearing bags), achieving state-of-the-art results, in contrast to classical approaches for gait recognition that use hand-crafted features, mainly based on binary silhouettes or dense tracklets.
In terms of classification strategies, an ensemble of \OVA linear SVM is a good choice, although a NN approach on PCA compressed descriptors offers similar accuracy, not requiring any training step.
Finally, we have shown that our automatically learnt gait signatures are suitable for gender recognition, what would allow to filter out a set of individuals before running a finer identification procedure.
As future work, we plan to extend our study to other datasets for gait recognition where multiple viewpoints are available and other CNN architectures combining OF with RGB data. 
{\small
\bibliographystyle{ieee}
\bibliography{shortstrings,local,bibAVA}
}

\end{document}